\documentclass[10pt,twocolumn,letterpaper]{article}

\usepackage{cvpr}
\usepackage{times}
\usepackage{epsfig}
\usepackage{graphicx}
\usepackage{amsmath}
\usepackage{amssymb}

\usepackage{relsize}
\usepackage{url}
\usepackage{tabularx, booktabs}
\usepackage{pifont}
\usepackage{float}
\usepackage[table]{xcolor}
\definecolor{lightgray}{gray}{0.9}
\definecolor{lightblue}{rgb}{0.93,0.95,1.0}
\definecolor{darkgreen}{rgb}{0.0,0.6,0.0}
\definecolor{mypink1}{rgb}{0.858, 0.188, 0.478}

\newcommand{\ignore}[1]{}

\newcolumntype{L}[1]{>{\raggedright\let\newline\\\arraybackslash\hspace{0pt}}m{#1}}
\newcolumntype{C}[1]{>{\centering\let\newline\\\arraybackslash\hspace{0pt}}m{#1}}
\newcolumntype{R}[1]{>{\raggedleft\let\newline\\\arraybackslash\hspace{0pt}}m{#1}}
\renewcommand{\eqref}[1]{Eq.~\ref{#1}}

\newcommand{\figref}[1]{Fig.~\ref{#1}}
\newcommand{\tabref}[1]{Tab.~\ref{#1}}
\newcommand{\secref}[1]{Sec.~\ref{#1}}

\newcommand{\reals}{\mathbb{R}}

\def\beq{\begin{equation}}
\def\eeq{\end{equation}}

\def\beqary{\begin{eqnarray}}
\def\eeqary{\end{eqnarray}}

\def\beqarz{\begin{eqnarray*}}
\def\eeqarz{\end{eqnarray*}}

\usepackage[pagebackref=true,breaklinks=true,letterpaper=true,colorlinks,bookmarks=false]{hyperref}

\cvprfinalcopy 

\def\cvprPaperID{0024} 

\ifcvprfinal\fi

\usepackage{boxedminipage}
\usepackage{bbold}

\renewcommand{\xi}{{\xx}^{(m)}}

\newcommand{\needcite}[1]{}

\newcommand{\be}{\begin{equation}}
\newcommand{\ee}{\end{equation}}
\newcommand{\benn}{\begin{equation*}}
\newcommand{\eenn}{\end{equation*}}
\newcommand{\bea}{\begin{eqnarray*}}
\newcommand{\eea}{\end{eqnarray*}}
\newcommand{\bean}{\begin{eqnarray}}
\newcommand{\eean}{\end{eqnarray}}

\newcommand{\xx}{\boldsymbol{x}}

\newcommand{\comment}[1]{}

\newcommand{\polyring}[1]{\reals\left[x_1,\ldots,x_n\right]}

\definecolor{atomictangerine}{rgb}{0.8, 0.2, 0.1}
\definecolor{turq}{rgb}{0.0, 0.5, 0.5}
\definecolor{darkturq}{rgb}{0.0, 0.4, 0.4}
\definecolor{bright}{rgb}{0.8, 0.1, 0}
\definecolor{darkgray}{gray}{0.3}
\definecolor{mahogany}{rgb}{0.6, 0.05, 0.05}
\definecolor{pink}{rgb}{1,0.05,0.6}
\definecolor{myblue}{rgb}{0.3,0.05,0.9}

\renewcommand{\eqref}[1]{Eq.~\ref{#1}}

\begin{document}

\title{Accurate Visual Localization for Automotive Applications}

\author{
Eli Brosh$^{^\star}$, \,\,
Matan Friedmann$^{^\star}$, \,\,
Ilan Kadar$^{^\star}$, \,\,
Lev Yitzhak Lavy$^{^\star}$, \,\,
Elad Levi$^{^\star}$, \,\, 
Shmuel Rippa$^{^\star}$, \,\,\\
Yair  Lempert, \,\, 
Bruno Fernandez-Ruiz, \,\,
Roei Herzig, \,\,
Trevor Darrell \vspace{3pt}\\
Nexar Inc.}


\maketitle

\renewcommand*{\thefootnote}{$\star$}
\setcounter{footnote}{1}
\footnotetext{Equal Contribution.}
\renewcommand*{\thefootnote}{\arabic{footnote}}
\setcounter{footnote}{0}
\thispagestyle{empty}


\begin{abstract}
\label{sec:abstract}
Accurate vehicle localization is a crucial step towards building effective Vehicle-to-Vehicle networks and automotive applications. Yet standard grade GPS data, such as that provided by mobile phones, is often noisy and exhibits significant localization errors in many urban areas. Approaches for accurate localization from imagery often rely on structure-based techniques, and thus are limited in scale and are expensive to compute. In this paper, we present a scalable visual localization approach geared for real-time performance.
We propose a hybrid coarse-to-fine approach that leverages visual and GPS location cues. Our solution uses a self-supervised approach to learn a compact road image representation. This representation enables efficient visual retrieval and provides coarse localization cues, which are fused with  vehicle ego-motion to obtain  high accuracy location estimates. 
As a benchmark to evaluate the performance of our visual localization approach, we introduce a new large-scale driving dataset based on video and GPS data obtained from a large-scale network of connected dash-cams. Our experiments confirm that our approach is highly effective in challenging urban environments, reducing localization error by an order of magnitude.

\end{abstract}
\section{Introduction}
\label{sec:introduction}

\begin{figure}[t!]
	\begin{center}
        \includegraphics[width=\linewidth]{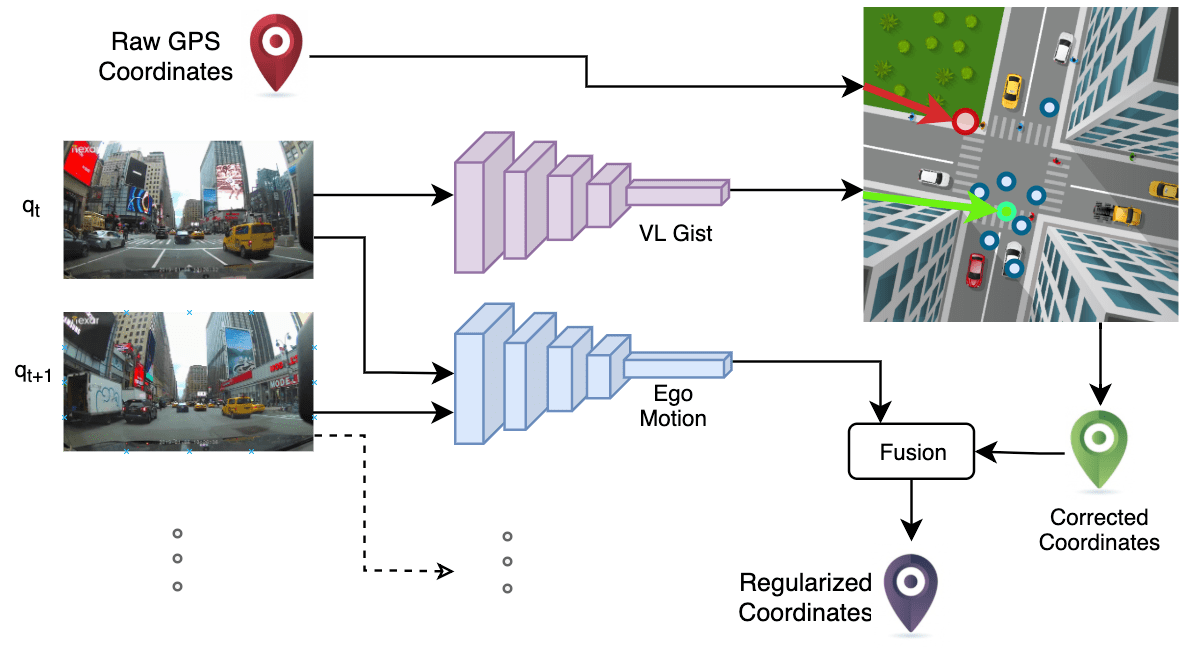}
	\caption{Method Overview: Given a video stream of images, a hybrid visual search and ego-motion approach is applied to leverage both image representation and temporal information. The VL-GIST representation is applied to provide a coarse localization fix of the image, while the visual ego-motion is used to to estimate the vehicle's motion between consecutive video images. Fusing vehicle dynamics with the coarse location fixes further regularizes the localization error and yields a high accuracy location data stream.
}
	\label{fig:method_overview}
	\end{center}
\end{figure}

\begin{figure}[ht!]
	\begin{center}
        \includegraphics[width=\linewidth]{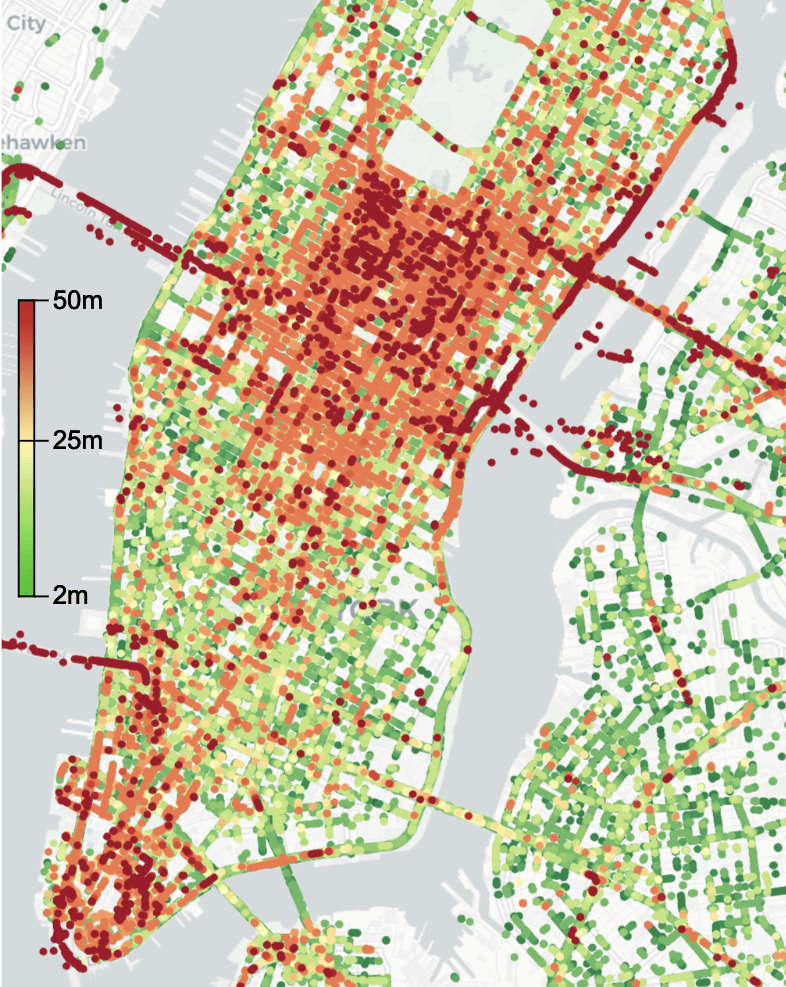}
	\caption{Accuracy of GPS data crowd-sourced from over 250K driving hours in NYC. The percentage of rides that experience GPS errors of 10 meters or more (likely due to urban canyons effects) is 40\%, and that of 20 meters or more is 20\%.}
	\label{fig:gps_accuracy_analysis_nyc}
	\end{center}
\end{figure}

Robust and accurate vehicle localization plays a key role in building safety applications based on Vehicle-to-Vehicle (V2V) networks. 
A V2V network allows vehicles to communicate with each other and to share their location and state, thus creating a 360-degree 'awareness' of other vehicles in proximity that goes beyond the line of sight.
According to the National Highway Traffic Safety Administration (NHTS), such a V2V network offers the promise to significantly reduce crashes, fatalities, and improve traffic congestion \cite{V2VNHTSA}.   
The increasingly ubiquitous presence of smartphones and dashcams,  with  embedded GPS and camera sensors as well as efficient data connectivity, provides an opportunity to implement a cost-effective V2V ''Ground Traffic Control Network''. Such a platform would facilitate cooperative collision avoidance by providing advance V2V warnings, e.g., intersection movement assist to warn a driver when it is not safe to enter an intersection due to high collision probability with other vehicles.
While GPS is widely used for navigation systems, its localization accuracy poses a critical challenge for proper operation of V2V safety networks. In some areas such as urban canyons environments,  GPS signals  are often blocked or partially available due to high-rise buildings \cite{noms_2016}. In \figref{fig:gps_accuracy_analysis_nyc} we show the accuracy of GPS readings from crowd-sourced data of over 250K driving hours taken in New York City (NYC). The figure demonstrates that the number of rides that suffer from urban canyon effects resulting in GPS errors of 10\,m or above is 40\%, and that of 20 meters is 20\%.    

In this work, we propose a hybrid coarse-to-fine approach for accurate vehicle localization in urban environments based on visual and GPS cues. \figref{fig:method_overview} shows a high level overview of the proposed solution\footnote{Part of Figure 1 was designed by macrovector/Freepik.}. 
First, a self-supervised approach is applied on a large-scale driving dataset to learn a compact representation, called \textit{Visual-Localization-GIST (VL-GIST)}. The representation preserves the geo-location distances between road images to facilitate robust and efficient coarse image-based localization. 
Then, given a driving  video stream, a hybrid visual search and ego-motion approach is applied by matching the extracted descriptor in the low embedded space against a restricted set of relevant geo-tagged images to provide a coarse localization fix; the coarse fix is fused with the vehicle ego-motion to regularize localization errors and obtain a high accuracy location stream.
  
To evaluate our model on realistic driving data, we introduce a challenging dataset based on real-world dashcam and GPS data.  We collect millions of images from more than 5 million rides, focusing on the area of NYC. 
Our experimental results show that an efficient visual search with the VL-GIST descriptor can reduce a mobile phone's GPS location error from 50 meters (often measured in urban areas) to under 10 meters, and that incorporating visual ego-motion further reduces the error to below 5 meters. 

Our contributions are summarized as follows:
\begin{itemize}
  \item We perform large-scale analysis of GPS quality in urban areas, and generate a comprehensive dataset for benchmarking vehicle localization in such areas (Sec.~\ref{sec:datasets}).
  \item We introduce a scalable approach for accurate and efficient localization that is geared for real-time performance  (Sec.~\ref{sec:method}).
  \item We conduct extensive evaluation of our approach in challenging urban environments and demonstrate an order of magnitude reduction in localization error (Sec.~\ref{sec:experiment}).
\end{itemize}

\section{Related Work}
\label{sec:related_work}

%

\textbf{SfM and Visual Ego-Motion.} 
The Structure from Motion (SfM) approach (e.g., \cite{Schnberger2015FromSI}) uses a 3D scene model of the world constructed from the geometrical relationship of overlapping images. For a given query image, 2D-3D correspondences are established using descriptor matching (e.g., SIFT \cite{Lowe_2004_IJCV}). These matches are then used to estimate the camera pose. This approach is not always robust, especially when the query images are taken under significantly different conditions compared to the database images, or on straight roads that are not close to intersections and do not have enough perpendicular visual queues; the computational demands of this method mean it is not presently feasible to scale to millions of cars.
Visual ego-motion, or visual odometry, is a well studied topic \cite{Scaramuzza2011VisualO}. Traditional methods use
 a complex pipeline including many steps such as feature extraction, feature
matching, motion estimation, local optimisation, etc which require a great deal of manual tuning. Early attempts of solving this problem using deep learning techniques still involved complex additional steps such as computing dense optical flow \cite{Costante2016ExploringRL} or using SfM to label the data \cite{Kendall2015ConvolutionalNF} to work. Wang at al \cite{Wang2017DeepVOTE} were the first to suggest an end-to-end approach using a recurrent neural network and show competitive performance to state-of-the-art methods. Other directions use stereo images \cite{Zhan_2018_CVPR, Li2018UnDeepVOMV}, an approach that is not viable to our setup.

\textbf{Retrieval Approaches}
Many approaches use image retrieval techniques to find the most relevant database images for each query image \cite{state_of_the_art_survey, state_of_the_art_survey_2018}. These assume that a database of geo-tagged reference images is provided. Given this database, they estimate the position of a new query image by searching for a matching image from the database. The leading methods for image retrieval operate by constructing a vector, called descriptor, constructed in such a way that the  distance between descriptors of similar images is smaller than the distance between descriptors corresponding to distinct images. All descriptors of a large database of images are recorded to a data base. To locate similar image to a query image we compute it's descriptor and then get a ranked list of images from the data base ordered by descriptors distances. Since the descriptors are often vectors of high dimension, a common practice is to apply a dimensionality reduction step of using PCA with whitening followed by L2-normalization \cite{ECCV12}. The evolution of descriptors for image retrieval problems are summarized in the survey paper of Zheng et al. \cite{Zheng2018SIFTMC}.
In urban areas this problem is particularly difficult due to repetitive structures \cite{Torii:2015:VPR:2881666.2882191,Jgou2009OnTB}, changes over time because of change of seasons, day and night and change in the construction elements \cite{Torii2015247PR} and the existence of many dynamic objects that are not related to the landmark that is being searched for, like vehicles.

\textbf{Traditional Descriptors.} Conventional image retrieval techniques rely on aggregation of local descriptors with methods based on "bag-of-word" representations \cite{Sivic:2003:VGT:946247.946751}, vectors of locally aggregated descriptors (VLAD) \cite{Jegou:2012:ALI:2360767.2361217}, Fisher vectors \cite{Perronnin2010LargescaleIR} and/or GIST \cite{Douze:2009:EGD:1646396.1646421}. The practical image retrieval task is composed of an initial filtering task where the descriptors in the database are ranked according to their distance to the descriptor of the query image and a second re-ranking phase which refines the ranking, using local descriptors, so to reduce ambiguities and bad matches. Such methods include query expansion \cite{QueryExpantion1, QueryExpantion2,QueryExpantion3} and spatial matching \cite{Philbin2007ObjectRW, Shen2012ObjectRA}.

\textbf{Descriptor Learning.} In the last few years convolutions neural networks (CNN) proved to be a powerful image representation for various recognition tasks so several authors have proposed the use of the activations of convolutional layers as local features that can be aggregated into a descriptor  suitable for image retrieval \cite{Babenko2015AggregatingLD,Razavian:2014:CFO:2679599.2679731}. However such approaches are not compatible with the geometric-aware models involved in the final re-ranking stages and thus can not compete with the state-of-the-art methods.
%
%
Since we want that the distance between two descriptors of similar images will be smaller than the distance between descriptor of two distinct images, it is natural to consider network architectures  developed for metric learning such as siamese   \cite{RTC16} or triplet  \cite{Schroff2015FaceNetAU, Wang2014LearningFI} learning networks. 
Arandjelovi\'c et al \cite{Arandjelovic16} propose a new training layer, NetVLAD, that can be plugged in any CNN architecture. The architecture mimics the classical approaches, that is local descriptors are extracted and then pooled in an
orderless manner to finally produce a fixed size unit descriptor. A dataset for training the network was constructed by using the Google Street View Time
allowing accessing multiple street-level panoramic
images taken at different times at close-by spatial locations. The authors demonstrated that NetVlad descriptor outperformed state-of-the-art learned and not-learned descriptors on the Pittsburgh 250k \cite{Torii2013VisualPR} and the Tokyo 24/7 \cite{Torii2015247PR} datasets. A further step of dimensionality reduction using PCA with whitening followed by L2-normalization  \cite{ECCV12} is applied to reduce the large NetVLAD, namely 16k or 32k,  descriptor to a size of 4096.
The R-MAC network of Tiolias et al. \cite{Tolias16} was develop to allow applying geometric aware methods for re-ranking and it does so by producing a global image
representation by aggregating the activation features of a CNN in a fixed layout
of spatial regions, followed by whitening with PCA. The descriptor produced by R-MAC is of compact, between 256 and 512,  dimension. Gordo at al. \cite{Gordo2016DeepIR, Gordo:2017} proposed using a triplet loss to train the R-MAC architecture  and a block for learning the pooling mechanism of the R-MAC descriptor. The network was trained on a large public dataset \cite{Babenko2014NeuralCF}. The dataset is very noisy and thus geometric filtering with SIFT keypoint detection were used to find positive examples. The authors demonstrated that this descriptor outperforms global descriptors and more complex systems deploying geometric verification and keypoint matching.
Radenovi{\'c} et al \cite{RTC16, Radenovic2018FinetuningCI} proposed using a siamease network with the contrastive loss. The positive and negative examples are selected in an unsupervised manner, by clustering a large collection of unlabeled images, using state-of-the-art SfM system \cite{Schnberger2015FromSI}. Since SfM system use strict geometrical verification procedures, the 3D models reliably guide the selection of matching and non-matching pairs.  
Zho et al. \cite{AttentionPyrVisPlaceRecog18} proposed and attention-based pyramid aggregation network (APANet) consisting of a spatial pyramid pooling block, attention block and  sum pooling block. They also proposed a fully unsupervised dimensioanlity and whitenings solution referred to as power PCA. The dataset used for training is the same as for NetVLAD \cite{Arandjelovic16}.

\section{Datasets}
\label{sec:datasets}
{\bf Data collection.}  Our data was collected from a large-scale deployment of connected dashcams. Each vehicle is equipped with a dashacam and a companion smartphone app that continuously captures and uploads sensor data such as GPS readings. Overall, the vehicles collected more than 5 million rides in the NYC area. From these rides, we use more than 200 million images for the image similarity dataset and more than 1000 video sequences for the ego motion dataset\footnote{The publicly available dataset can be found at: \href{https://github.com/getnexar/Nexar-Visual-Localization}{https://github.com/getnexar/Nexar-Visual-Localization}}. 

\subsection{Image Similarity Dataset} \label{subsec:imageSim}
\begin{figure}[t!]
	\begin{center}
        \includegraphics[width=\linewidth]{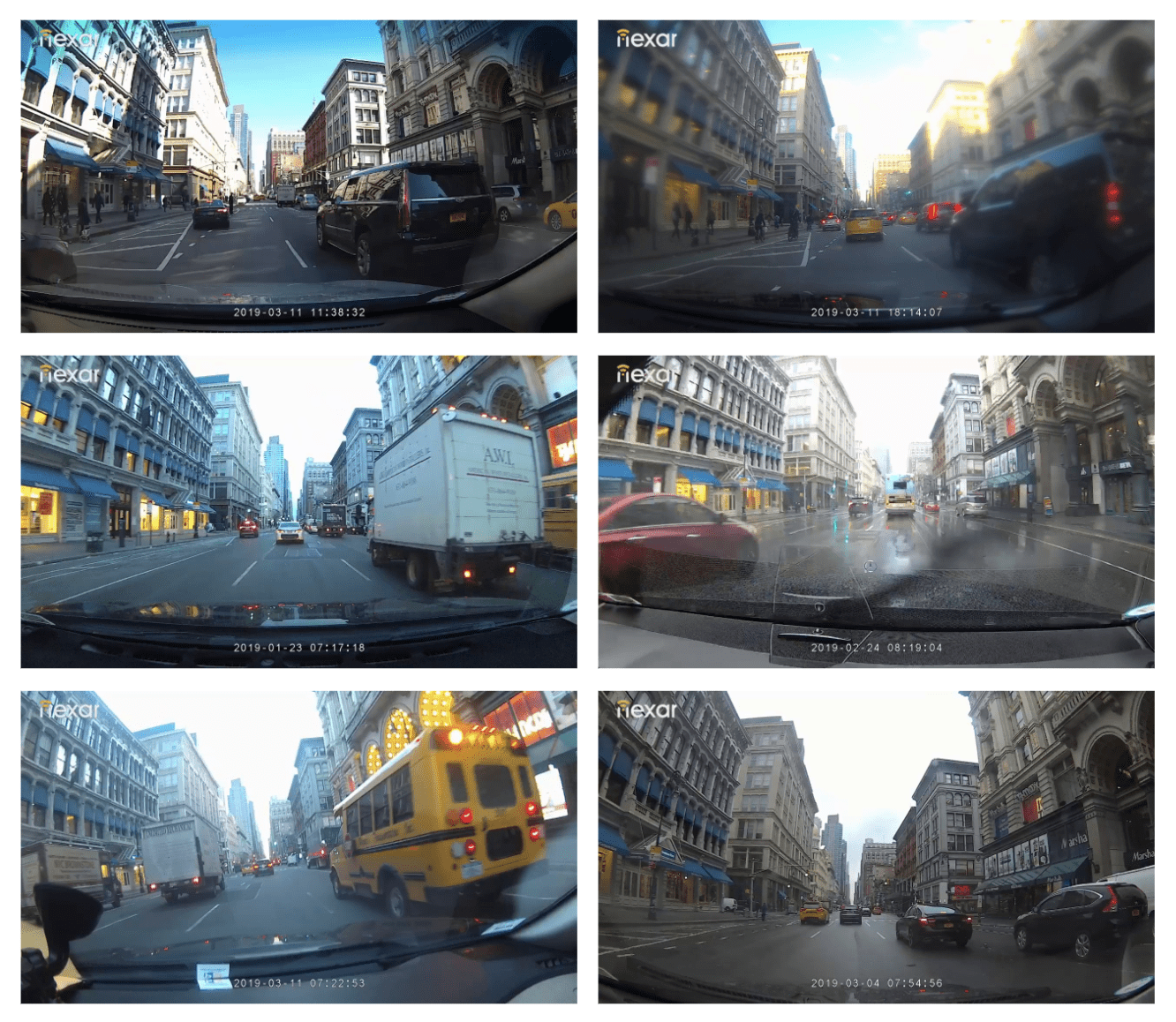}
	\caption{Example images captured from the same cell. Each square cell of 10x10 meters consist of large-set of images acquired with different weather and lighting conditions as well as different dynamic objects, e.g., vehicles or pedestrians. This dataset allows generating models that are invariant to weather, lightning, dynamic objects and addition or removal of construction elements. } 
	\label{fig:many_images_on_grid}
	\end{center}
\end{figure}

From the complete image similarity dataset we collect a subset of geo-tagged images for which the reported accuracy of the GPS signal is better than 10 meters. We found that at nearly all places in NYC, excluding tunnels, we have enough images with the required GPS accuracy. In fact, at least 10\% of the collected images have the required accuracy. Thus each square cell of 10x10 meters contains many images acquired with different weather and lighting conditions, different dynamic objects, e.g. vehicles or pedestrians, and different time, as demonstrated in  \figref{fig:many_images_on_grid}. This dataset allows generating models that are invariant to weather, lightning, dynamic objects and addition or removal of construction elements. 

Images taken by dash-cams are frames taken from a video. Each video, in turn is a part of a full ride of a single vehicle. Since there is a large correlation between images of a single video or ride, we save also the ride ID that is further used for triplet sampling.

The dataset is organized in a spatial data-structure that allows fast access to neighbouring images of each image in the data where we interpret neighbouring relation as being close geographically, and also in orientation.

\subsection{Video Dataset with Sub-Meter Location Accuracy}
In order obtain a benchmark with accurate location sequences of meter-level accuracy, we created a route annotation tool, which shows side-by-side the route on an aerial imagery (as a series of raw GPS points) and the corresponding driving video. A human annotator can align the ground view video with the overhead (aerial) view, and then correct the location of route points accordingly. 

Since this is a complex annotation task, we generated a test set for annotators and selected the top 3 experts. By checking the consistency of the route corrections across the different annotations, we observe a localization error with a mean of one meter in urban areas, and up to four meters on highways. 

\begin{figure}[t!]
	\begin{center}
        \includegraphics[width=\linewidth]{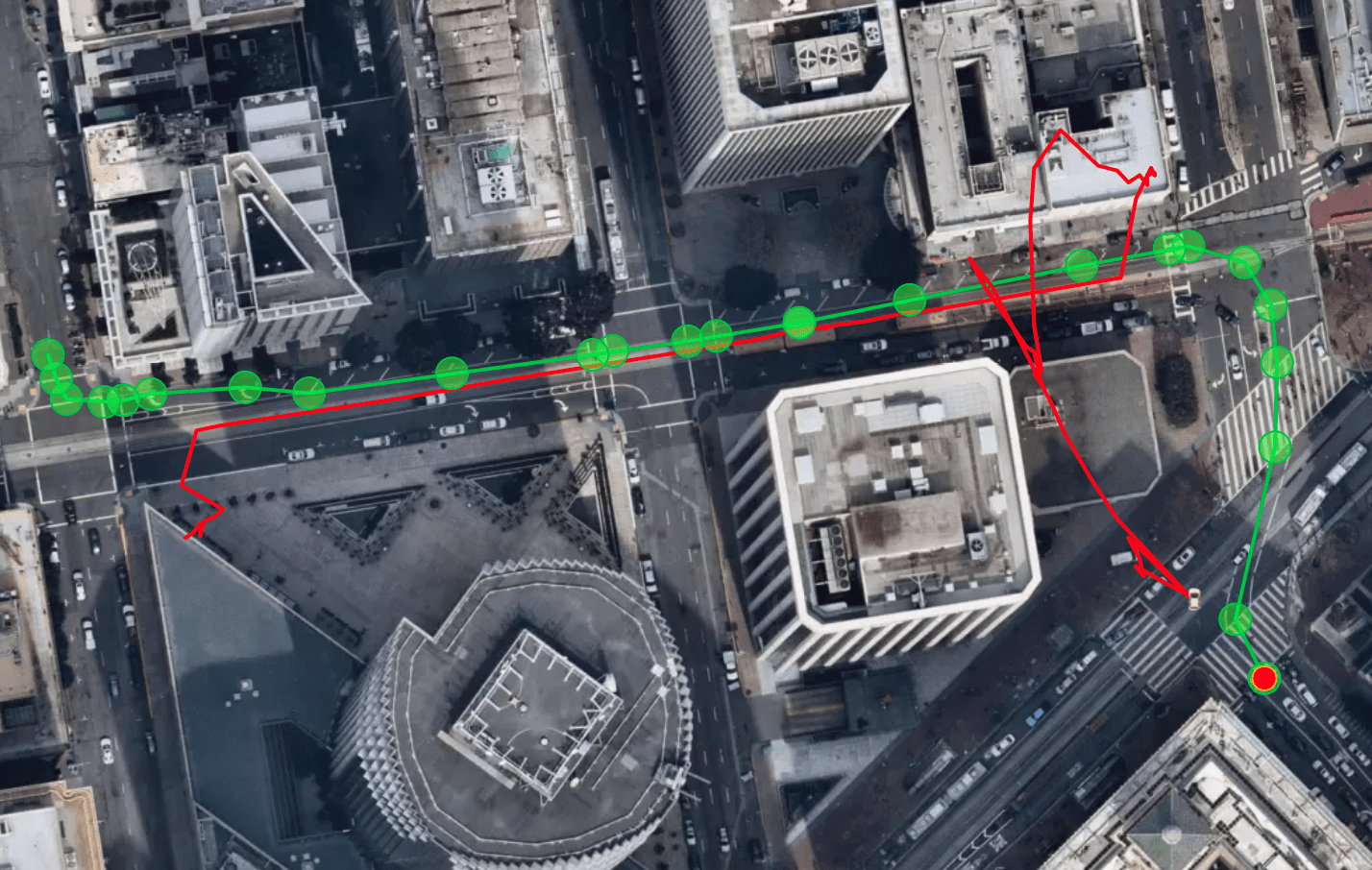}
	\caption{We created a route annotation tool which facilities  location corrections by aligning a route on a aerial map with a corresponding driving video. The image shows a comparison between the manually annotated location series (green dots) and the raw GPS data (red route). } 
	\label{fig:gps_vs_annotated_gt}
	\end{center}
\end{figure}

\figref{fig:gps_vs_annotated_gt} shows an example ride with a comparison between the manually annotated location series and the raw GPS data.

\section{Method}
\label{sec:method}

We improve the raw location data using a hybrid approach consisting of visual similarity to obtain coarse location fixes (i.e., of 10 meter accuracy) and further refinement and regularization using visual ego-motion to yield an accurate location stream (i.e., of 5 meter accuracy). 

\subsection{Self-Supervised Learning from Triplets}
The model structure is a deep CNN followed by three small fully connected layers where the final layer is $L_2$ normalized. The network is trained in a self-supervised manner with a variant of the triplet loss: Let $f(x) \in \mathcal{R}^d$ be the output of the embedding layer for an image $x$ and let $x^a, x^p, x^n$ be the anchor, positive and negative images. Then the triplet loss is just the cross entropy loss of
\[
\textup{softmax}( (D_p,D_n))
\]
where $D_p = ||f(x^a)-f(x^p)||_2$, the positive distance, is the distance between the embedding of the anchor image and the embedding of the positive image and  $D_n = ||f(x^a)-f(x^n)||$ is the negative distance. 

\begin{figure}[ht!]
	\begin{center}
        \includegraphics[width=\linewidth]{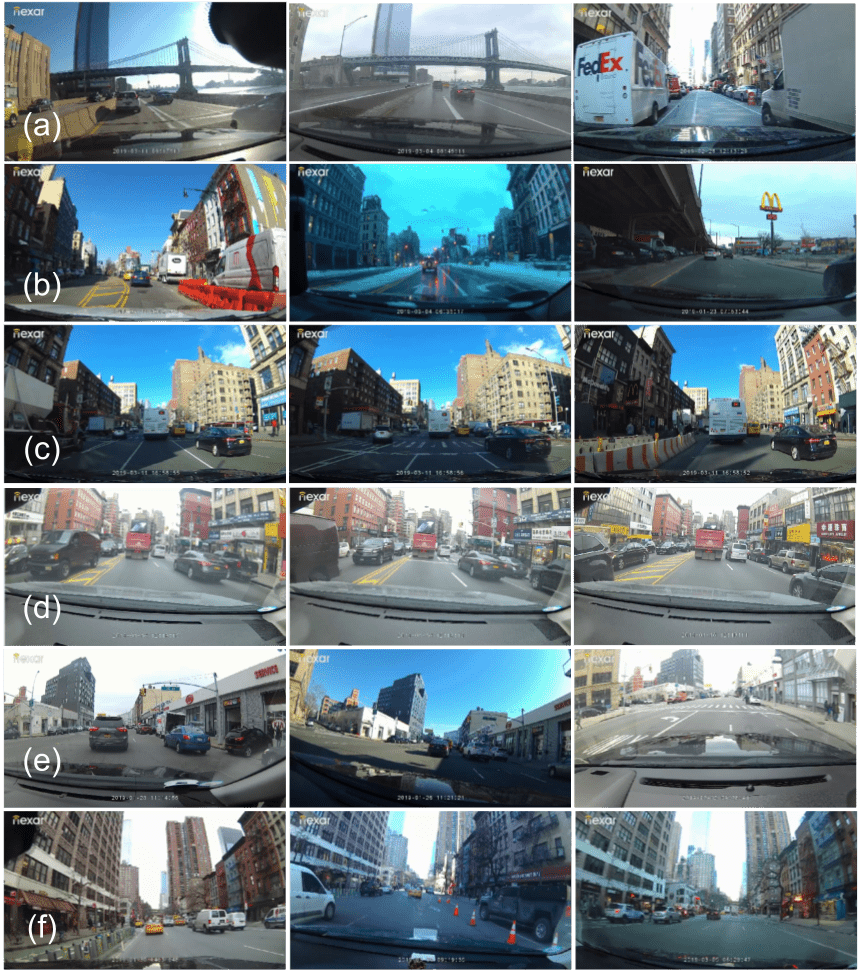}
	\caption{Examples of the three types of triplets. In each row, the leftmost two images are matching in location and heading, while the rightmost frame is the negative example of that triplet. (a) Regular triplets showing the Brooklyn Bridge from two different rides compared with a randomly sampled street. (b) Another regular triplet example, showing invariance to weather and lighting conditions. (c)+(d) Ride triplet showing two close frames and one negative frame from the same ride. (e)+(f) Hard negative triplet showing invariance to lighting conditions and and camera orientation.}
	\label{fig:triplets}
	\end{center}
\end{figure}

In order to effectively train the model with the triplet loss to produce a good embedding, we utilize our image similarity dataset as a source for our triplet sampling. In particular we produce three triplet generators:
\begin{itemize}
	\item \textbf{Regular triplets}. This generator produces triples in which the anchor image is close to the positive image and far from the negative image. The anchor image is randomly sampled from our dataset, the positive image is sampled from all images that are close up to 10 meters to the anchor image and are oriented in the same direction (namely, the difference in the GPS heading of the two images is up to 20 degrees) while the negative image is sampled from all images that are far away, say more than 500 meters from the anchor image. Special care is taken to assure than none of the images are from the same driving video. Two examples of  triplets sampled by this sampler are shown in \figref{fig:triplets} (a)-(b).
	\item \textbf{Random hard negative triplets}. This generator produces anchor and positive images in the same way as the regular triplet sampler but with harder negatives. More precisely, the negative image is sampled from images that are at distance between 20 to 30 meters from the anchor and roughly in the same orientation. Examples for such triples are shown in  \figref{fig:triplets} (e)-(f).
	\item \textbf{Video sampler.} We utilize the inherent spatial ordering between consecutive images in a video. First, we sample a video from a collection of driving videos and then sample an image from this video as an anchor. The positive image is the closest frame to the anchor provided that it's distance from the anchor is less than 10 meters. The negative image is the closest image to the anchor provided that it's distance from the anchor is between 25 and 50 meters. As before, a triplet is selected only if the anchor, positive and negative images have roughly the same orientation. While this sampling procedure generate triplets that are highly correlated it is still useful, on top of the other samplers, since we have high confidence in the spatial ordering and the relevancy of the negative example as shown in \figref{fig:triplets} (c)-(d).
\end{itemize}
During training we randomly sample one of the above generators and use it to produce the next triplet to train. This sampling methods guarantees that the embedding layer will be invariant to weather, illumination and dynamic objects such as vehicles or pedestrians. The video sampler and random hard negative samplers refine the embedding so that the descriptors produced reflect the notion of distance to the query image.

\subsection{Efficient Retrieval Inference}
\label{sec:method-retrieval}
The visual retrieval task boils down to comparing the descriptor of the query image to the database to obtain a ranked list of images form the database sorted by descriptors distances. A weighted average of the GPS coordinates of the k'th closest images, in descriptor space, yield a corrected GPS signal for the query image.

There are several factors contributing to the performance of our retrieval pipeline: Restricting the number of images in the database to be ranked, speeding up the ranking procedure by using small descriptors and eliminating the need for additional re-ranking procedures.

First, we have a geo-tagged image and thus we do not need to search the whole database for matching images. Thus we restrict our search only in an area of modest size around the query image according to the GPS accuracy. Because we rank images only in a small proximity to the query image, we discovered that we do not need any sort of re-ranking technique. The efficiency of the ranking procedure increases as the dimension of the descriptor decreases. We use a very simple triplet network \cite{Schroff2015FaceNetAU}, namely a deep CNN, followed by $L_2$ normalization and three fully connected layers that produce a small, 30 dimensional, embedding vector. This is in contrast to existing  methods, see \cite{Magliani2018AnAR} which compares many methods, that report on descriptor dimensions in the range between 128 and 32k. 



\subsection{Visual Ego-Motion Estimation}
\label{sec:method-egomotion}
The visual retrieval approach provides coarse localization fixes with a noise distribution, as captured by the confidence of location prediction. We use visual ego-motion to reduce this noise term. That is, we estimate the vehicle's motion between consecutive video frames, and fuse the vehicle dynamics with the coarse fixes to regularize the location coordinates, yielding a (high-rate) data stream with lower localization error.

\begin{figure}[t!]
	\begin{center}
        \includegraphics[width=\linewidth]{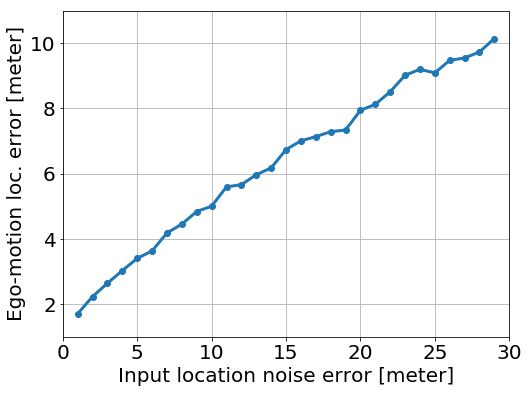}
	\caption{Localization error of the ego-motion prediction as a function of input location noise error in meters. These measurements were done by adding normally distributed noise to the ground truth at varying standard deviations, applying ego-motion, and extracting the regularized coordinates' error estimation. Using ego-motion yields a 2x-3x improvement in localization error. 
	}
	\label{fig:fusion_vs_noise_err}
	\end{center}
\end{figure}

{\bf Vehicle model.} We follow Ackerman's steering model~\cite{Musleh2012VisualEM} and capture the kinematic motion of the vehicle between two time steps by two parameters: (a) a rotation, occurring around the center motion of the rear part of the vehicle and (b) a forward translation after the rotation.

We use an end-to-end learning approach for ego-motion estimation, shown by recent work to be robust to image anomalies and imperfections~\cite{Costante2016ExploringRL}. We train a deep neural network, composed of CNN based feature extraction, that observes a sequences of images and aims to predict the motion of the vehicle. 
It takes as an input a monocular image sequence. At each time step, the two frames are resized, stacked together, and fed into the CNN to produce an effective feature for ego-motion estimation.  The convolution layers are followed by two dense layers, and then split to two heads. Each head is  composed of a 100 dimensional dense layer connected to a one dimensional dense layer.  The network is trained using accurate location supervisory sequences (see ~\secref{sec:datasets}) with a combined loss: Let $x$ be the stacked images, and let $t$ and $r$ be the corresponding ground truth values of translation and rotation. The loss term is then defined as \[
\frac{1}{2}|f^t(x)-t| + \frac{1}{2}|f^r(x)-r|
\]
where $f^t(x) \in \mathcal{R}$ and  $f^r(x) \in \mathcal{R}$ are the predicted translation and rotation values. We minimize the mean of the loss term across the whole training dataset. 

To compute the confidence of the ego-motion estimation, we split the values range of each ego motion parameter into multiple bins, and estimate the probability of
a parameter to fall within a bin. Aggregating bin values  around the mean yields an error range for the ego-motion predictions.

\subsection{Fusion Algorithm} We use a Kalman filter to compute  high accuracy location predictions. The state of the filter represents the 2D location of the vehicle in a cartesian coordinate system. The measurement inputs are the speed (translation divided by the inter-frame time) and steering of the vehicle, as computed by our ego-motion model; and the coarse 2D pose fixes from visual retrieval, each input with its noise estimation.  With each new ego-motion estimation, we modify the vehicle's 2D location according to the new rotation and translation values. When a new coarse pose measurement is available, we fuse it with the current state to compute an updated  location along with its uncertainty. Our Kalman filter formulation is similar to that found in Section III-B of~\cite{Pink2009Visual} with minor adjustments: we replace the pose measurements from the map matching (in~\cite{Pink2009Visual}) by pose measurements from visual retrieval (\secref{sec:method-retrieval}), and the measurements from visual odometry by those from ego-motion (\secref{sec:method-egomotion}).

\section{Experiments}
\label{sec:experiment}

\subsection{Implementation Details}
{\bf Visual retrieval model details.} We selected a ResNet50 \cite{He2016DeepRL} backbone and trained the network using 
the SGD optimizer with the 1cycle policy procedure described in \cite{Smith2017SuperConvergenceVF, Smith2018Asystematic} with a maximal learning rate of 0.003, minimum momentum of 0.85, maximum momentum of 0.95 and weight decay of 1e-6. 

To predict the location of an image, we first set a threshold by looking at the distribution of the distances in the VL-GIST feature space, of all the image tuples in the validation set which their location is less than 10 meter (we remove outlier samples). After getting the threshold we then predict the location of the queried image by a weighted average of the location of all the key-frames, which their distance in the image VL-GIST feature space to the quarried, is smaller than the threshold. The weights are determine by the ratio between the key-frame distance and the sum of the distances in the feature space. We predict the location only in cases where there are at least 5 neighbors that passed the threshold. We extract also a confidence score according to the distribution of the location of the neighbors.

\begin{figure}[t!]
	\begin{center}
        \includegraphics[width=\linewidth]{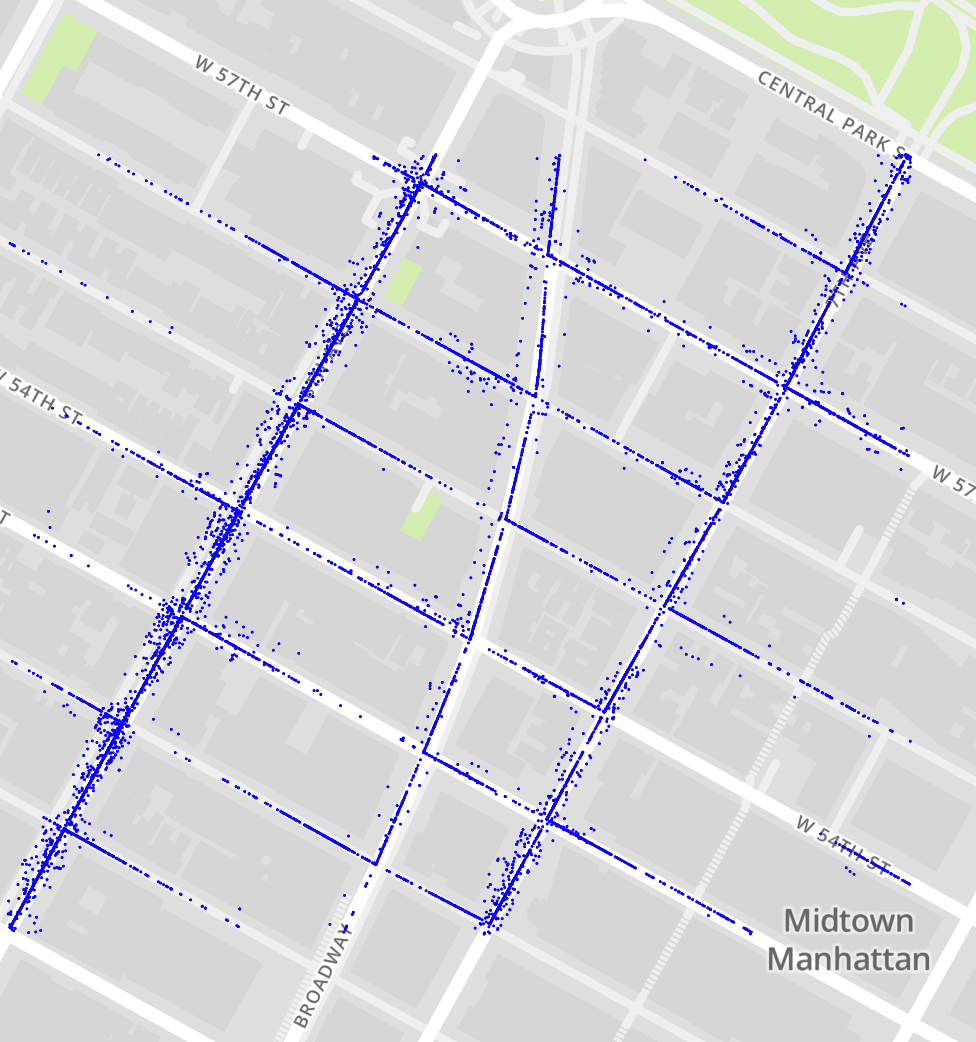}
	\caption{Distribution of the location of the key-frames in the test area. Key-frames were chosen to cover the area with an approximately uniform distribution along the drivable paths to avoid biases.} 
	\vspace{-10pt}
	\label{fig:validation_density}
	\vspace{-10pt}
	\end{center}
\end{figure}

{\bf Ego-motion model details.} To obtain an efficient implementation geared for running on mobile devices, we use a simple 8-layer CNN configuration with 2x2 fixed size filters  and a layer depth sequence of $[20, 30, 40, 60, 80, 120, 160, 240]$. We train the model using an SGD optimizer with a learning rate of 0.001 and a momentum of 0.9.  We use 1000 driving videos (from roughly 1000 different vehicles) as training set and 100 videos as test set.  Each video is approximately 40 seconds in length and has a resolution of $1280 \times 720$. We train the ego-motion model with two consecutive frames, each resized to 256x256. The frames are taken at various time intervals ranging from 33\,ms to 1\,sec. The approach not only significantly augments the training data but also enables the model to support dynamic infer rates, e.g., reducing computation overhead for static scenes when the vehicle is idle. 

\subsection{Evaluation Methodology and Results}

\begin{table}[t!]
\begin{center}
\label{results}
\small
  \begin{tabular}{|l|ccc|c|c|} 
     \hline
     \multicolumn{1}{|c|}{} &    \multicolumn{3}{c|}{Accuracy} & \multicolumn{1}{c|}{ME} & \multicolumn{1}{c|}{Recall}\\
      {} & {\textless 5m} & {\textless 10m} & {\textless 15m}  & {} & {}\\
     \hline
     \texttt{GPS-NN} & 0.09 & 0.24 & 0.39 & 21.5m & \textbf{0.97}\\
     \hline
     \texttt{VL-GIST} & 0.20 & 0.41 & 0.61 & 13.5m & 0.48\\
     \hline
     \textbf{VL-GIST*} & \textbf{0.30} & \textbf{0.63} & \textbf{0.82} & \textbf{9.7m} & 0.52  \\
     \hline
\end{tabular}
  \caption{Comparison between the three methods with 50 meter max GPS error.}
  \vspace{-10pt}
  \label{sec:exper:results2}
  \vspace{-15pt}
\hspace{2.0cm}
\end{center}
\end{table}

\begin{table}[t!]
\begin{center}
\label{results2}
\small
  \begin{tabular}{|l|ccc|c|c|} 
     \hline
     \multicolumn{1}{|c|}{} &    \multicolumn{3}{c|}{Accuracy} & \multicolumn{1}{c|}{ME} & \multicolumn{1}{c|}{Recall}\\
      {} & {\textless 5m} & {\textless 10m} & {\textless 15m}  & {} & {}\\
     \hline
     \texttt{GPS-NN} & 0 & 0.01 & 0.02 & 82.7m & \textbf{1}\\ 
     \hline
     \texttt{VL-GIST} & 0.12 & 0.32 & 0.48 & 23.1m & 0.42\\
     \hline
     \textbf{VL-GIST*} & \textbf{0.23} & \textbf{0.52} & \textbf{0.74} & \textbf{15.4m} & 0.41  \\
     \hline
\end{tabular}
  \caption{Comparison between the three methods with 200 meter max GPS  error.}
  \label{sec:exper:results3}
  \vspace{-25pt}
\hspace{2.0cm}
\end{center}
\end{table}

\subsubsection{Visual retrieval for coarse localization}
To estimate the visual localization quality we select an area of $750 \times 280$ square meters from the Image similarity dataset (see ~\secref{subsec:imageSim}). We hold out all the images from the test area (i.e., the triplet network was not trained on images from this area). We call these images key-frames (see ~\figref{fig:validation_density}).

We set a maximal GPS error threshold (varies between 50-200 meter according to the experiment). For each key-frame, we randomly distorted the GPS location up to the maximal GPS error. Then we predict the location of the image, according to its VL-GIST nearest neighbors in the radius of the maximal GPS error, and compare it the the GPS location of the image.

We compare the features extracted from the triplet network (VL-GIST) and the triplet network with the refinement triplets (VL-GIST*). Since image locations are not evenly distributed in our data, we also compare against naive baseline approach, called GPS-NN, of averaging the 10 nearest neighbors with respect to the geo-location distance. 

For each method we compare between the percentage of the errors that were less than 5,10 and 15 meters. We also compare the mean error and the recall rate for each method. As can be seen from ~\tabref{sec:exper:results2} and ~\tabref{sec:exper:results3}, even when looking at maximal error of 50 meter, the road VL-GIST distance preform much better comparing to the geographical distance.
The experiments also demonstrate the value of training the networks with the refinements triplets and the affect on the final results.

\vspace{-10pt}
\subsubsection{Visual ego-motion for localization refinement}
To estimate the visual ego-motion refinement quality, we add to the Ground-Truth (GT) location a random noise with a normal distributions, where the standard deviation varies between 3 meter to 30 meter. We then fixed the distorted location using the ego-motion and estimated the mean error relative to the GT.

Running this test on various location noise values, as can be seen in ~\figref{fig:fusion_vs_noise_err}, we find that within an acceptable error range of up to 30 meters, the ego-motion fusion correction yields an approximate factor of 2-3 in improvement of the localization error.

\begin{figure}[t!]
	\begin{center}
        \includegraphics[width=\linewidth]{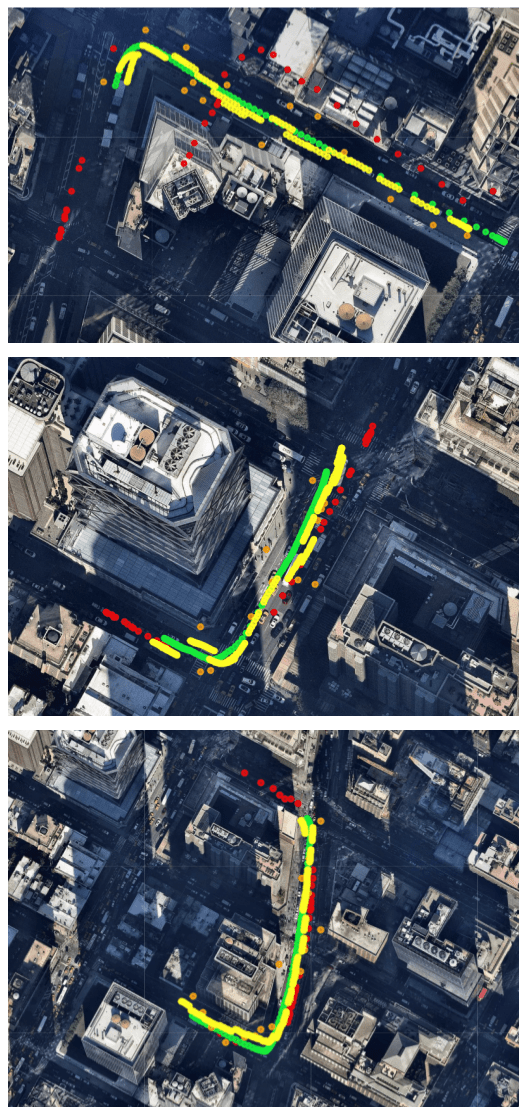}
	\caption{Visualization of the entire process on three example rides. Green dots show the ground truth coordinates, red dots show the raw GPS coordinates, orange dots show the VL-GIST prediction, and yellow dots show the regularized final coordinates. } 
	\label{fig:final_waypoints}
	\vspace{-20pt}
	\end{center}
\end{figure}

\begin{figure}[t!]
	\begin{center}
        \includegraphics[width=\linewidth]{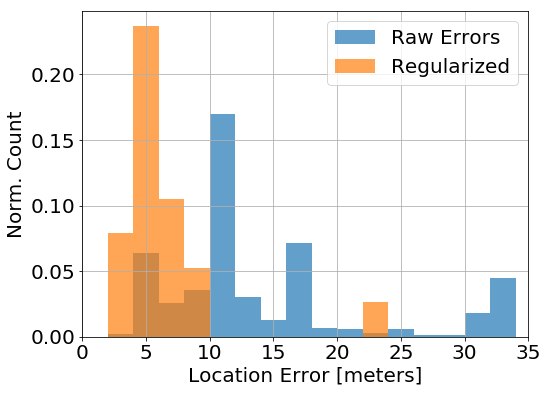}
	\caption{Comparison of the normalized distributions of the raw and regularized localization errors. The raw reported errors (blue) are aggregated from 250K different rides, and are spread out over a wide range, with under 1\% beyond the 35m error range. After regularizing the coordinates by fusing VL-GIST coarse correction with the ego-motion output, the distribution of localization errors becomes much more compact and can be approximated to a normal distribution around 5 meters.} 
	\label{fig:gps_err_dist_comp}
	\vspace{-20pt}
	\end{center}
\end{figure}

Moreover, we compare in \figref{fig:gps_err_dist_comp} the original raw GPS coordinates' error distribution with the localization error distribution of the regularized coordinates, when combining the results from both the visual retrieval component and the ego-motion component. The normalized distributions show that we were able to reduce the variance in the localization error, and lower the mean error to be distributed compactly around 5 meters. 

\figref{fig:final_waypoints} shows the visualization of the entire process on three example rides in NYC.

\vspace{-10pt}
\section{Conclusion}
\vspace{-5pt}
\label{sec:summary and future work}

In this work, we address the challenge of vehicle localization and a propose a scalable approach for accurate and efficient visual localization geared for real time performance.

We first perform a large-scale analysis of GPS quality in urban areas, and generate comprehensive dataset for benchmarking vehicle localization in these areas.  
We then introduce a hybrid coarse-to-fine approach for accurate vehicle localization in urban environments based on efficient visual search and ego-motion. A low-dimensional global descriptor is introduced for fast retrieval of coarse localization, which is then fused with the vehicle ego-motion to regularize localization error and to provide high accuracy localization stream.
Next, we introduce a large-scale dataset based on real-world dashcam and GPS data to evaluate our model on realistic driving data.
Finally, we conduct an extensive evaluation of our approach in challenging urban environments and demonstrate a order of magnitude reduction in localization error.

In future work we would like to explore improvements in the method's efficiency by reducing the dimension of the VL-GIST descriptor. For that, we can utilize our triplet sampling policy within any triplet architecture suggested for deep hashing (e.g.,  \cite{Norouzi2012HammingDM, Wang2016DeepSH, Liu2018DeepTQ}). In addition, we would like to study the relationship between the localization performance and the amount of visual data that is used for learning the VL-GIST representation.

{\small
\bibliographystyle{ieee}
\bibliography{main}
}

\end{document}